\documentclass[runningheads]{llncs}

\usepackage{times}
\usepackage{epsfig}
\usepackage{graphicx}
\usepackage{amsmath}
\usepackage{amssymb}
\usepackage{color}
\usepackage{subcaption}
\usepackage{booktabs}
\usepackage{wrapfig,lipsum,booktabs}
\usepackage[font=small,labelfont=bf]{caption}
\usepackage[]{algorithm2e}

\usepackage[width=122mm,left=12mm,paperwidth=146mm,height=193mm,top=12mm,paperheight=217mm]{geometry}

\usepackage{bm} 
\newcommand{\vct}[1]{{#1}} 
\newcommand{\mat}[1]{{#1}} 
\newcommand{\cst}[1]{{#1}}  
\newcommand{\eat}[1]{}

\newcommand{\vp}{\vct{p}}
\newcommand{\vx}{{\vct{x}}}
\newcommand{\vy}{\vct{y}}

\newcommand{\vv}{\vct{v}}
\newcommand{\vw}{\vct{w}}

\newcommand{\cN}{\cst{N}}
\newcommand{\cM}{\cst{M}}

\newcommand{\cD}{\cst{D}}

\newcommand{\cT}{\cst{T}}

\newcommand{\sY}{\mathcal{Y}}
\newcommand{\sV}{\mathcal{V}}
\newcommand{\calY}{\mathcal{Y}}
\newcommand{\mL}{\mat{L}}
\newcommand{\mI}{\mat{I}}
\newcommand{\mK}{\mat{K}}

\newcommand{\vpi}{\vct{\pi}}
\newcommand{\mOmega}{\mat{\Omega}}

\newcommand{\ours}{Seq\textit{G}DPP}
\newcommand{\gdpp}{\textit{G}DPP}

\usepackage[pagebackref=true,breaklinks=true,letterpaper=true,colorlinks,bookmarks=false]{hyperref}

\begin{document}

\pagestyle{headings}
\mainmatter
\def\ECCV18SubNumber{972}  

\title{Improving Sequential Determinantal Point Processes for Supervised Video Summarization} 

\title{Improved Sequential Determinantal Point Processes for Supervised Video Summarization} 

\titlerunning{Improving Sequential DPP for Supervised Video Summarization}

\authorrunning{A. Sharghi, A. Borji, C. Li, T. Yang, B. Gong}

\author{Aidean Sharghi$^1$, Ali Borji$^1$, Chengtao Li$^2$, Tianbao Yang$^3$, Boqing Gong$^4$\\
{\tt\small aidean.sharghi@knights.ucf.edu, ctli@mit.edu, tianbao-yang@uiowa.edu, boqinggo@outlook.com}
}

\institute{$^1$CRCV, University of Central Florida, Orlando, Florida\\
$^2$Massachusetts Institute of Technology, Cambridge, Massachusetts \\
$^3$University of Iowa, Iowa City, Iowa \\ $^4$Tencent AI Lab, Seattle, Washington\\}
\maketitle

\begin{abstract}
   
It is now much easier than ever before to produce videos. While the ubiquitous video data is a great source for information discovery and extraction, the computational challenges are unparalleled. Automatically summarizing the videos has become a substantial need for browsing, searching, and indexing visual content. This paper is in the vein of supervised video summarization using sequential determinantal point process (SeqDPP), which models diversity by a probabilistic distribution. We improve this model in two folds. In terms of learning, we propose a large-margin algorithm to address the exposure bias problem in SeqDPP. In terms of modeling, we design a new probabilistic distribution such that, when it is integrated into SeqDPP, the resulting model accepts user input about the expected length of the summary. Moreover, we also significantly extend a popular video summarization dataset by 1) more egocentric videos, 2) dense user annotations, and 3) a refined evaluation scheme. We conduct extensive experiments on this dataset (about 60 hours of videos in total) and compare our approach to several competitive baselines.

\end{abstract}

\section{Introduction}
It is now much easier than ever before to produce videos due to ubiquitous acquisition capabilities. The videos captured by UAVs and drones, from ground surveillance, and by body-worn cameras are easily reaching the scale of gigabytes per day. In 2017, it was estimated that there were at least 2.32 billion active camera phones in the world~\cite{obile2016ericsson}. In 2015, 2.4 million GoPro body cameras were sold worldwide~\cite{hirsch2017seizing}. While the big video data is a great source for information discovery and extraction, the computational challenges are unparalleled. Automatically summarizing the videos has become a substantial need for browsing, searching, and indexing visual content. 

Under the \textit{extractive} video summarization framework, a summary is composed of important shots of the underlying video. This notion of importance, however, varies drastically from work to work in the literature. Wolf defines the importance as a function of motion cues~\cite{wolf1996key}. Zhao and Xing formulate it by reconstruction errors~\cite{zhao2014quasi}. Gygli et al.\ learn a mixture of \textit{interestingness}, \textit{representativeness}, and \textit{uniformity} measures to find what is important~\cite{gygli2015video}. These differences highlight the complexity of video summarization. The criteria for summarizing vastly depend on the content, styles, lengths, etc.\ of the video and, perhaps more importantly, users' preferences. For instance, to summarize a surveillance video, a running action might flag an important event whereas in a football match it can be a normal action observed throughout the video.

To overcome those challenges, there are two broad categories of approaches in general. One is to reduce the problem domain to a homogeneous set of videos which share about the same characteristics (e.g., length and style) so that experts can engineer some domain-specific criteria of good summaries~\cite{sun2014ranking,potapov2014category}. The other is to design models that can learn the criteria automatically, often from human-annotated summaries in a supervised manner~\cite{gong2014diverse,sharghi2016query,sharghi2017query,zhang2016video,sadeghian2018automatic}. The latter is more appealing because a model can be trained for different settings of choice, while the former is not as scalable. 

This paper is also in the vein of supervised video summarization based on determinantal point process (DPP)~\cite{kulesza2012determinantal}. Arising from quantum physics and random matrix theories, DPP is a powerful tool to balance importance and diversity, two axiomatic properties in extractive video summarization. Indeed, a good summary must be collectively diverse in the sense that it should not have redundancy of information.  Moreover, a shot selected into the summary must add value to the quality of the summary; otherwise, it is not important in the context of the summary. Thanks to the versatility of DPP and one of its extension called SeqDPP~\cite{gong2014diverse} for handling sequences, they have been employed in a rich line of recent works on video summarization~\cite{sharghi2016query,sharghi2017query}. 

This paper makes two-pronged contribution towards  improving these models to more effectively learn better video summarizers. In terms of learning, we propose a large-margin algorithm to address the SeqDPP's exposure bias problem explained below. In terms of modeling, we design a new probabilistic block such that, when it is integrated into SeqDPP, the resulting model accepts user input about the expected length of the summary.


We first explain the exposure bias problem with the existing SeqDPP works --- it is actually a mismatch issue in many sequence to sequence (seq2seq) learning methods~\cite{sutskever2011generating,sutskever2014sequence,bahdanau2014neural,vinyals2015grammar,serban2016building}. When the model is trained by maximizing the likelihood of user annotations, the model takes as input  user annotated ``oracle'' summaries. At the test time, however, the model generates output by searching over the output space in a greedy fashion and its intermediate conditional distributions may receive input from the previous time step that deviates from the oracle. In other words, the model is exposed to different environments in the training and testing stages, respectively. This exposure bias also results in the loss-evaluation mismatch~\cite{ranzato2015sequence} between the training phase and the inference. To tackle these issues, we adapt the \textit{Large-Margin} algorithm originally derived for training LSTMs~\cite{wiseman2016sequence} to the SeqDPPs. The main idea is to alleviate the {exposure bias} by incorporating inference techniques of the test time into the objective function used for training. Meanwhile, we add to the large-margin formulation a multiplicative reward term that is directly related to the evaluation metrics to mitigate the loss-evaluation mismatch.

In addition to the new large-margin learning algorithm, we also improve the SeqDPP model by a novel probabilistic distribution in order to allow users to control the lengths of system-generated video summaries. To this end, we propose a generalized DPP (\textit{G}DPP) in which an arbitrary prior distribution can be imposed over the sizes of subsets of video shots. As a result, both vanilla DPP and $k$-DPP~\cite{DBLP:conf/icml/KuleszaT11} can be considered as special instances of \textit{G}DPP. Moreover, we can conveniently substitute the (conditional) DPPs in SeqDPP by \textit{G}DPP. When a user gives an expected length of the summary, we dynamically allocate it to different segments of the video and then choose the right numbers of video shots from corresponding segments.

We conduct extensive experiments to verify the improved techniques for supervised video summarization. First of all, we significantly extend the UTE dataset~\cite{lee2012discovering} and its annotations of video summaries and per-shot concepts~\cite{sharghi2017query} by another eight egocentric videos~\cite{fathi2012social}. Following the protocol described in~\cite{sharghi2017query}, we collect three user summaries for each of the hours-long videos as well as concept annotations for each video shot. We evaluate the large-margin learning algorithm on not only the proposed sequential \textit{G}DPP but also the existing SeqDPP models. 

\section{Related work and background}
We briefly review the related work in this section. Besides, we also describe the major body of DPPs and SeqDPPs. Readers are referred to~\cite{kulesza2012determinantal} and \cite{gong2014diverse} for more details and properties of the two versatile probability models.

\paragraph{Supervised video summarization.} In recent years, data-driven learning to tackle research problems has attracted plenty of attention. This is mainly because they can learn complex relations from data, specially when the underlying relations are unknown. Video summarization is an instance of such cases. The fact that different users prefer different summaries is a strong evidence to complexity of the problem. To overcome the impediments, one solution is to learn how to make good summaries in a supervised manner. The degree of supervision, however, is different in the literature. In~\cite{khosla2013large,kim2014joint,xiong2014detecting,chu2015video}, weakly supervised web image and video priors help define visual importance, captions associated with videos used by~\cite{song2015tvsum,liu2015multi} to infer semantic importance. Finally, many frameworks (e.g.,~\cite{zhang2016video,gong2014diverse,sharghi2016query,sharghi2017query,gygli2015video}) learn a summarizer directly from user-annotated summaries.

\paragraph{Sequence-to-Sequence Learning.} Sequence-to-sequence (Seq2seq) modeling has been successfully employed in a vast set of applications, especially in Natural Language Processing (NLP). By the use of Recurrent Neural Networks (RNNs), impressive modeling capabilities and results are achieved in various fields such as machine translation~\cite{bahdanau2014neural} and text generation applications (e.g., for image and video captioning~\cite{venugopalan2015sequence,xu2015show}).

Seq2seq models are conveniently trained as conditional language models, maximizing the probability of observing next ground truth word conditioned on the input and target words. This translates to using merely a word-level loss (usually a simple cross-entropy over the vocabulary). 

While the training procedure described above has shown to be effective in various word-generation tasks, the learned models are not used as conditional models during inference at test time. Conventionally, a greedy approach is taken to generate the output sequence. Moreover, when evaluating, the complete output sequence is compared against the gold target sequence using a sequence-level evaluation metric such as ROUGE~\cite{lin2004rouge} and BLEU~\cite{papineni2002bleu}.

\paragraph{Determinantal point process (DPP).} A discrete DPP~\cite{kulesza2012determinantal,hough2006determinantal} defines a distribution over all the subsets of a  ground set measuring the negative correlation, or repulsion, of the elements in each subset. Given a ground set $\mathcal{Y}=\{1,...,N\}$, one can define $\mat{K}\in\mathbb{R}^{N\times N}$, a positive semi-definite kernel matrix that represents the per-element importance as well as the pairwise similarities between the $N$ elements. A distribution over a random subset $Y\subseteq\mathcal{Y}$ is a DPP, if for every $\vct{y}\subseteq\mathcal{Y}$ the following holds:
\begin{equation}
    P(\vct{y}\subseteq\mathcal{Y};\mat{K}) = \det({\mat{K}_{\vct{y}}})
    \label{ePy}
\end{equation}
where $\mat{K}_{\vct{y}}$ is the squared sub-kernel of $\mat{K}$ with rows and columns indexed by the elements in $\vct{y}$, and $\det(.)$ is the determinant function. $\mat{K}$ is referred to as the marginal kernel since one can compute the probability of any subset $\vct{y}$ being included in $\mathcal{Y}$. It is the property of the determinant that promotes diversity: in order to have a high probability $P(i,j \in Y; \mat{K})= \mat{K}_{ii}\mat{K}_{jj} - \mat{K}_{ij}^2$, the per-element importance terms $\mat{K}_{ii}$ and $\mat{K}_{jj}$ must be high and meanwhile the pairwise similarity terms $K_{ij}$ must be low. 

To directly specify the atomic probabilities for all the subsets of $\mathcal{Y}$, Borodin and Rains derived another form of DPPs through a positive semi-definite matrix $\mat{L}=\mat{K}(\mat{I}-\mat{K})^{-1}$~\cite{borodin2005eynard}, where $\mat{I}$ is an identity matrix. It samples a subset $\vct{y} \subseteq \mathcal{Y}$ with probability
\begin{equation}
P_{\mat{L}}(Y= \vct{y} ; \mat{L}) = \frac{\det(\mat{L}_{\vct{y}})}{\det(\mat{L}+\mat{I})},
\label{eDPPLensemble}
\end{equation}
where the denominator $\det(\mat{L}+\mat{I})$ is a normalization constant. 

\eat{
\paragraph{$k$-DPP.} Conditioning on a fixed size $k$ of sampled subsets, the resulting distribution is named $k$-DPPs~\cite{DBLP:conf/icml/KuleszaT11}. Denote by $\mat{L}=\sum_{n=1}^{N}\lambda_n\vct{v}_n\vct{v}_n^T$ the eigen-decomposition of the DPP kernel. For any subset $\vct{y}\subseteq\mathcal{Y}$ and $|\vct{y}|=k$, a $k$-DPP can be expressed as
\begin{align}
P_k(Y=\vct{y}) := P_L(Y=\vct{y}||\vct{y}|=k)= \frac{\det(\mat{L}_{\vct{y}})}{\sum_{|\vct{z}|=k}\det(\mat{L}_{\vct{z}})} = \frac{\det(\mat{L}_{\vct{y}})}{\sum_{|\vct{z}|=k} \prod_{n\in \vct{z}}\lambda_n}. \nonumber
\end{align}
}

\paragraph{Sequential DPP (seqDPP).}\label{sSeqDPP}
Gong et al.\
proposed SeqDPP~\cite{DBLP:conf/nips/GongCGS14} to preserve partial orders of the elements in the ground set. Given a long sequence $\mathcal{V}$ of elements (e.g., video shots), we divide them into $\mathsf{T}$ disjoint yet consecutive partitions $\bigcup_{t=1}^\mathsf{T} {\mathcal{V}}_t = \mathcal{V}$. The elements within each partition are orderless to apply DPP and yet the orders among the partitions are observed in the following manner. At the $t$-th time step, SeqDPP selects a diverse subset of elements by a variable $X_t\subseteq{\mathcal{V}}_t$ from the corresponding partition and conditioned on the elements $\vct{x}_{t-1}\subseteq{\mathcal{V}}_{t-1}$ selected from the previous partition. In particular, the distribution of the subset selection variable $X_t$ is given by a conditional DPP,
\begin{align}
    P(X_t=\vct{x}_{t} | X_{t-1}=\vct{x}_{t-1}) :=& P_L(Y_t=\vct{x}_{t}\cup\vct{x}_{t-1} | \vct{x}_{t-1}\subseteq{Y_t}; \mat{L}^t) \\
    = &P_L(X_t=\vct{x}_t; \mat{\Omega}^{t}) = \frac{\det{\mat{\Omega}^t_{\vct{x}_{t}}}}{\det(\mat{\Omega}^t+\mat{I})}, \label{eConditionalDPP}
\end{align}
where $P_L(Y_t;\mat{L}^t)$ and $P_L(X_t;\mat{\Omega}^t)$ are two L-ensemble DPPs with the ground sets $\vct{x}_{t-1}\cup\mathcal{V}_t$ and $\mathcal{V}_t$, respectively --- namely, the conditional DPP itself is a valid DPP over the ``shrinked'' ground set. The relationship between the two L-ensemble kernels $\mat{L}^t$ and $\mat{\Omega}^t$ is given by~\cite{borodin2005eynard},
\begin{align}
\mat{\Omega}^{t} = \left([(\mat{L}^t+\mat{I}_{\mathcal{V}_t})^{-1}]_{\mathcal{V}_t}\right)^{-1}-\mat{I}, \label{eNewL}
\end{align}
where $\mat{I}_{\mathcal{V}_t}$ is an identity matrix of the same size as $\mat{L}^t$ except that the diagonal entries corresponding to $\vct{x}_{t-1}$ are 0's, $[\cdot]_{\mathcal{V}_t}$ is the squared submatrix of $[\cdot]$ indexed by the elements in $\mathcal{V}_t$, and the number of rows/columns of the last identity matrix $\mat{I}$ equals the size of the $t$-th video segment $\mathcal{V}_t$.

\section{A large-margin algorithm for learning SeqDPPs}
We present the main large-margin learning algorithm in this section. We first review the mismatch between the training and inference of SeqDPPs~\cite{gong2014diverse} and then describe the large-margin algorithm in detail.

\paragraph{Training and inference of SeqDPP.} For the application of supervised video summarization, SeqDPP is trained by maximizing the likelihood (MLE) of user summaries. At the test time, however, an approximate online inference is employed: 
\begin{align}
\label{eInf}
    \hat{\vct{x}}_1 = \text{argmax}_{\vct{x}\in \mathcal{V}_1} P(X_1 = \hat{\vct{x}}),\quad
    \hat{\vct{x}}_2 = \text{argmax}_{\vct{x}\in \mathcal{V}_2} P(X_2 = \hat{\vct{x}}|X_1=\hat{\vct{x}}_1), \quad 
    ...
\end{align}
We note that, in the inference phase, a possible error at one time step (e.g., $\hat{\vct{x}}_1$) propagates to the future but MLE always feeds the oracle summary to SeqDPP in the training stage (i.e., exposure bias~\cite{ranzato2015sequence}). Besides, the likelihood based objective function used in training does not necessarily correlate well with the evaluation metrics in the test stage (i.e., loss-evaluation mismatch~\cite{ranzato2015sequence}).  

The issues above are common in seq2seq learning. It has been shown that improved results can be achieved if one tackles them explicitly~\cite{daume2009search,ross2011reduction,collins2004incremental,ranzato2015sequence,shen2015minimum}. Motivated by thثse findings, we propose a large-margin algorithm for SeqDPP to mitigate the exposure bias and loss-evaluation mismatch issues in existing SeqDPP works. Our algorithm mainly borrows some ideas from \cite{wiseman2016sequence}, which studies the large-margin principle in training recurrent neural networks. However, we are not constrained by the beam search, do not need to change the probabilistic SeqDPP model to any non-probabilistic version, and also fit a test-time evaluation metric into the large-margin formulation.

We now design a loss function
\begin{align}
    \mathcal{L}(\theta) = \sum_{t=1}^T \delta(x_{1:t-1}^*\cup \hat{x}_t,x_{1:t}^*) M(x_t^*,\hat{x}_t,x_{t-1}^*;\mat{L}), 
\end{align}
which includes two components: 1) a sequence-level cost $\delta$ which allows us to scale the loss function depending on how erroneous the test-time inference is compared to the oracle summary, and 2) a margin-sensitive loss term $M$ which penalizes the situation when the probability of an oracle sequence fails to exceed the probability of the model-inferred ones by a margin. Denote by $\hat{x}_t$ and $\hat{x}_{t}^*$ the subsets selected from the $t$-th partition $\mathcal{V}_t$ by SeqDPP and by an ``oracle'' user, respectively. Let $x_{1:t}^*$ represent the oracle summary \textit{until} time step $t$. The sequence-level cost $\delta(x_{1:t-1}^* \cup \hat{x}_t,x_{1:t}^*)$ can be any metric (e.g., $1-\text{F-score}$) used to contrast a system-generated summary with a user summary.

Assuming SeqDPP is able to choose the right subset $x_{t-1}^*$ from partition $\mathcal{V}_{t-1}$, given the next partition $\mathcal{V}_t$, the margin-sensitive loss penalizes the situation that the model selects a different subset $\hat{x}_t$ from the oracle $x_t^*$,
\begin{align}
M(x_t^*,\hat{x}_t,x_{t-1}^*;\mat{L}) :=& [1 - \log P(X_t=x_t^*|x_{t-1}^*) + \log P(X_t=\hat{x}_t|x_{t-1}^*)]_+ \notag\\
= &[1-\log \det(\mat{L}_{x_t^* \cup x_{t-1}^*})+\log \det(\mat{L}_{\hat{x}_t \cup x_{t-1}^*})]_+
\end{align}
where $[\cdot]_+=\max(\cdot,0)$. When we use this loss term in training SeqDPP, we always assume that the correct subset $\hat{x}_{t-1}=x_{t-1}^*$ is chosen at the previous time step $t-1$. In other words, we penalize the model step by step instead of checking the whole sequence of subsets predicted by the model. This allows more effective training by 1) forcing the model to choose correct subsets at every time step, and 2) enabling us to set the gradient weights according to how erroneous a mistake at this time step actually is in the eyes of evaluation metric.

\eat{
Moreover, to directly optimize the model for the evaluation metric, we define $\delta(.)$ as:

\begin{equation}
    \delta (y_1,y_2) = 1 - g(y_1,y_2)
\end{equation}

where $g$ is the evaluation metric that measures a sequence $y_1$ against the target sequence $y_2$. This allows using virtually \textbf{any} evaluation metric, possibly a \textbf{non-differentiable} function, directly in the training to solve the \textit{Loss-Evaluation Mismatch}. Furthermore, by using the model's own prediction, model is successfully exposed to its own errors and is forced to update its parameters accordingly, addressing the \textit{Exposure Bias}.

To minimize $\mathcal{L}(\theta)$, we compute the gradients using REINFORCE algorithm~\cite{sutton1998reinforcement,williams1992simple}. The expected gradient is approximated using a sample from the model, however, we set the sample to be the test-time inference output of the model using current parameters:

\begin{equation}
    \triangledown_\theta \mathcal{L}(\theta) = -\sum_{t=1}^T \delta(\bar{x}_t,x_{1:t}) \triangledown_\theta M(x_t,x^{'}_t,x_{t-1},\mat{L})
\end{equation}
}

Compared to MLE, it is especially appealing that the large-margin formulation flexibly takes the evaluation metric into account. As a result, it does not require SeqDPP to predict exactly the same summaries as the oracles. Instead, when the predicted and oracle summaries are equivalent (not necessarily identical) according to the evaluation metric, the model parameters are not updated. 

\section{Disentangling size and content in SeqDPP} \label{sApproach}

In this section, we propose a sequential model of generalized DPPs (\ours) that accepts an arbitrary distribution over the sizes of the subsets whose content follow DPP distributions. It allows users to provide priors or constraints over the total items to be selected. We first present the generalized DPP and then describe how to use it to devise the sequential model, \ours.

\subsection{Generalized DPPs ({\gdpp}s)}
Kulesza and Taskar have made an intriguing observation about the vanilla DPP: it conflates the size and content of the variable $Y$ for selecting subsets from the ground set $\sY$~\cite{DBLP:conf/icml/KuleszaT11}. To see this point more clearly, we can re-write a DPP as a mixture of elementary DPPs $P_E(Y)$~\cite[Lemma 2.6]{DBLP:journals/corr/abs-1207-6083},
\begin{align}
P_L(Y;\mL) &= \frac{1}{\det(\mL+\mI)}\sum_{J\subseteq \mathcal{Y}} P_E(Y;J)\prod_{n\in J}\lambda_n,  \\
& \propto\sum_{k=0}^\cN\sum_{{J\subseteq \mathcal{Y}, |J|=k}} P_E(Y;J)\prod_{n\in J}\lambda_n  \label{eMixElem}  
\end{align}
where the first summation is over all the possible sizes of the subsets and the second is about the particular items of each subset. 

Eigen-decomposing the L-ensemble kernel to $\mL=\sum_{n=1}^{\cN}\lambda_n\vv_n\vv_n^T$, the marginal kernel of the elementary DPP $P_E(Y;J)$ is $\mK^J=\sum_{n\in J}\vv_n\vv_n^T$ --- it is interesting to note that, due to this form of the marginal kernel, the elementary DPPs do not have their counterpart L-ensembles. The elementary DPP $P_E(Y;J)$ always chooses $|J|$ items from the ground set $\calY$, namely, $P(|Y|=|J|)=1$. 

Eq.~(\ref{eMixElem}) indicates that, to sample from the vanilla DPP, one may sample the size of a subset from a uniform distribution followed by drawing items/content for the subset. We propose to perturb this process and explicitly impose a distribution $\vpi=\{\pi_k\}_{k=0}^\cN$ over the sizes of the subsets, \begin{align}
P_G(Y;\mL) \;\propto\; \sum_{k=0}^\cN \;\;\pi_k \sum_{{J\subseteq \sY, |J|=k}}\; P(Y;J)\prod_{n\in J}\lambda_n 
\label{eDefineGDPP}
\end{align}
As a result, the generalized DPP (\gdpp) $P_G(Y;\mL)$ entails both DPP and $k$-DPP~\cite{DBLP:conf/icml/KuleszaT11} as special cases (when $\vpi$ is uniform and when $\vpi$ is a Dirac delta distribution, respectively), offering a larger expressive spectrum. Another interesting result is for a truncated uniform distribution $\vpi$ over the sizes of the subsets. In this case, we arrive at a DPP which selects subsets with bounded cardinality, $P(Y\,|\,k_1\le |Y|\le k_2;\mL)$. Such constraint arises from real applications like
document summarization, image display, and sensor placement.

\paragraph{Normalization.}
The normalization constant for \gdpp is $Z_G = \sum_{{J\subseteq{\sY}}}\pi_{|J|} \prod_{n\in J}\lambda_n$. Details are included in the supplementary materials (Suppl.). The computation complexity of this normalization depends on the eigen-decomposition of $\mL$. With the eigenvalues $\lambda_n$, we can compute the constant $Z_G$ in polynomial time $O(\cN^2)$ with some slight change to the recursive algorithm~\cite[Algorithm 7]{DBLP:journals/corr/abs-1207-6083}, which calculates all the elementary symmetric polynomials $\sum_{{|J|=k}}\prod_{n\in J}\lambda_n$ for $k=0,\cdots,\cN$ in $O(\cN^2)$ time. Therefore, the overall complexity of computing the normalization constant for {\gdpp} is about the same as the complexity of normalizing an L-ensemble DPP (i.e., computing $\det(\mL+\mI)$).

\paragraph{Evaluation.} With the normalization constant $Z_G$, we are ready to write out the probability of selecting a particular subset $\vy\subseteq{\sY}$ from the ground set by \gdpp,
\begin{align}
P_G(Y=\vy;\mL) = \frac{\pi_{|\vy|}}{Z_G} \det(\mL_{\vy}) \label{eGDPP}
\end{align}
in which the concise form is due to the property of the elementary DPPs that $P_E(Y=\vy;J)=0$ when $|\vy|\neq |J|$. 

\paragraph{{\gdpp} as a mixture of $k$-DPPs.}
The {\gdpp} expressed above has a close connection to the $k$-DPPs~\cite{DBLP:conf/icml/KuleszaT11}. This is not surprising due to the definition of {\gdpp} (cf.\ Eq.~(\ref{eDefineGDPP})). Indeed, {\gdpp} can be exactly interpreted as a mixture of $\cN+1$ $k$-DPPs $P_k(Y=\vy;\mL), k=0,1,\cdots,\cN$,
\begin{align}
P_G(Y=\vy;\mL) = \frac{\pi_{|\vy|}\sum_{{|J|=|\vy|}}\prod_{n\in J}\lambda_n}{Z_G} P_{|\vy|}(Y=\vy;\mL) \notag
\end{align}
if \textbf{\em all the $k$-DPPs}, i.e., the mixture components, \textbf{\em share the same L-ensemble kernel $\mL$} as \gdpp. 

If we introduce a new notation for the mixture weights, $p_k \triangleq \pi_k/Z_G \sum_{{|J|=k}}\prod_{n\in J}\lambda_n$, the {\gdpp} can then be written as
\begin{align}
\label{eGDPPMix}    P_G(Y;\mL) = \sum_{k=0}^\cN p_k P_k(Y;\mL).
\end{align}
Moreover, there is no necessity to adhere to the involved expression of $p_k$. Under some scenarios, directly playing with $p_k$ may significantly ease the learning process. We will build a sequential model upon the {\gdpp} of form~(\ref{eGDPPMix}) in the next section.

\paragraph{Exact sampling.} Following the interpretation of {\gdpp} as a weighted combination of $k$-DPPs, we have the following decomposition of the probability:
\begin{align}
P(Y | Y\sim \text{\gdpp}) = P(Y | Y\sim \text{$k$-DPP}) P(k | k\sim \text{\gdpp}), \notag
\end{align}
where, with a slight abuse of notation, we let $k\sim\text{\gdpp}$ denote the probability of sampling a $k$-DPP from \gdpp. Therefore, we can employ a two-phase sampling procedure from the \gdpp,
\begin{itemize}
\item Sample $k$ from the discrete distribution $\vp=\left\{p_i\right\}_{i=0}^N$.
\item  Sample $Y$ from $k$-DPP.
\end{itemize}

\subsection{A sequential model of {\gdpp}s (\ours)}
In this section, we construct a sequential model of the generalized DPPs (\ours) such that not only it models the temporal and diverse properties as SeqDPP does, but also allows users to specify the prior or constraint over the length of the video summary.

We partition a long video sequence $\sV$ into $\cT$ disjoint yet consecutive short segments $\bigcup_{t=1}^\cT {\cal V}_t = \cal{V}$. The main idea of {\ours} is to adaptively distribute the expected length $\cM_0$ of the video summary to different video segments over each of which a {\gdpp} is defined. In particular, we replace the conditional DPPs in SeqDPP (cf.\ eq.~(\ref{eConditionalDPP})) by {\gdpp}s,
\begin{align}
    &P(X_t=\vx_{t} | X_{t-1}=\vx_{t-1}) \\ \triangleq & P_G(X_t=\vx_t; \mOmega^{t}) = {p^t_{|\vx_t|}} P_{|\vx_t|}(X_t=\vx_t; \mOmega^t),
\end{align}
where the last equality follows Eq.~(\ref{eGDPPMix}), and recall that the L-ensemble kernel $\mOmega^t$ encodes the dependencies on the video frames/shots selected from the immediate past segment $\vx_{t-1}\subseteq{\sV_{t-1}}$ (cf.\ Section~\ref{sSeqDPP}, Eq.~(\ref{eNewL})). The discrete distribution $\vp^t=\{p^t_k\}$ is over all the possible sizes of the subsets at time step $t$.

We update $\vp^t$ adaptively according to
\begin{align}
    p^t_k \propto \exp(- \alpha(k-\mu^t)^2 ),
\end{align}
where the mean $\mu^t\in [0,|\sV_t|]$ is our belief about how many items should be selected from the current video segment $\sV_t$ and the concentration factor $\alpha>0$ tunes the confidence of the belief. When $\alpha$ approaches infinity, the {\gdpp} $P_G(X_t;\mOmega^t)$ degenerates to $k$-DPP and chooses exactly $\mu^t$ items into the video summary. 

Our intuition for parameterizing the mean $\mu^t$ encompasses three pieces of information: the expected length $\cM_0$ over the overall video summary, number of items that have been selected into the summary up to the $t$-th time step, and the variety of the visual content in the current video segment $\sV_t$. Specifically, 
\begin{align}
\mu^t \triangleq \frac{\cM_0 - \sum_{t'=1}^{t-1}|\vx_{t'}|}{\cT - t + 1} + \vw^T\phi(\sV_{t})  \label{eMut}
\end{align}
where the first term is the average number of items to be selected from each of the remaining video segments to make up an overall summary of length $\cM_0$, the second term $\vw^T\phi(\sV_t)$ moves around the average number depending on the current video segment $\sV_t$, and $\phi(\cdot)$ extracts a feature vector from the segment. We learn $\vw$ from the training data --- user annotated video summaries and their underlying videos. We expect that a visually homogeneous video segment gives rise to negative $\vw^T\phi(\sV_t)$ such that less than the average number of items will be selected from it, and vice versa. 

\subsection{Learning and inference}

For the purpose of out-of-sample extension, we shall parameterize {\ours} in such a way that, at time step $t$, it conditions on the corresponding video segment $\sV_t$ and the selected shots $X_{t-1}=\vx_{t-1}$ from the immediate previous time step. We use a simple convex combination~\cite{DBLP:conf/icml/KuleszaT11}
of $\cD$ base {\gdpp}s whose kernels are predefined over the video for the parameterization. Concretely, at each time step $t$,
\begin{align}
\nonumber      P(X_t|\vx_{t-1},\sV_t)=& P_G(X_t; \mOmega^t,\sV_t) 
    \triangleq \sum_{i=1}^\cD \beta_i P_G(X_t;\mOmega^{t(i)},\sV_t) \\ =&\sum_{k=0}^{|\calY_t|}p_k^t \sum_{i=1}^\cD \beta_i P_k(X_t;\mOmega^{t(i)},\sV_t)
\end{align}
where the L-ensemble kernels $\mOmega^{t(i)}, i=1,\cdots,\cD$ of the base {\gdpp}s are derived from the corresponding kernels $\mL^{t(i)}$ of the conditional DPPs (eq.~(\ref{eNewL})). We compute different Gaussian RBF kernels for $\mL^{t(i)}$ from the segment $\sV_t$ and previously selected subset $\vx_{t-1}$ by varying the bandwidths. The combination coefficients ($\beta_i\geq 0, \sum_i\beta_i=1$) are learned from the training videos and summaries.

Consider a single training video $\sV=\cup_{t=1}^\cT\sV_t$ and its user summary $\{\vx_{t}\subseteq\sV_t\}_{t=1}^\cT$ for the convenience of presentation. We learn {\ours} by maximizing the log-likelihood,
\begin{align}
    &\mathcal{L} = \log \text{\ours} \nonumber = \sum_{t=1}^\cT\log P(X_t=\vx_t|\vx_{t-1},\sV_t)\\ 
    = &\sum_{t=1}^\cT \log p_{|\vx_t|}^t + \sum_{t=1}^\cT \log\bigg(\sum_{i=1}^\cD \beta_i P_{|\vx_{t}|}\Big(X_t=\vx_t;\Omega_i^{t(i)}\Big)\bigg).  \nonumber
\end{align}

\section{Experimental Setup and Results}

In this section, we provide details on compiling an egocentric video summarization dataset, annotation process, and the employed evaluation procedure.


\paragraph{Dataset.} While various video summarization datasets exist~\cite{gygli2014creating,song2015tvsum,de2011vsumm}, we put consumer grade egocentric videos in our priority. Due to their lengthy nature, they carry a high level of redundancy, making summarization a vital and challenging problem. UT Egocentric ~\cite{lee2012discovering} dataset includes 4 videos each between 3$\sim$5 hours long, covering activities such as driving, shopping, studying, etc. in an uncontrolled environment. However, we find this dataset insufficient for supervised video summarization summarization, hence, we significantly extend it by adding another 8 egocentric videos to it (averaging over 6 hours each) from social interactions dataset~\cite{fathi2012social}. These videos are recorded using head-mounted cameras worn by individuals during their visit to Disney parks. Our efforts results in a dataset consisting of 12 long videos with a total of over 60 hours of video content.


\paragraph{User Summary Collection.} Having compiled a set of 12 egocentric videos, we recruit three students to summarize the videos. The only instruction we give them is to operate on the 5-second video shot level. Namely, the full shot will be selected into the summary once any frame in the shot is chosen. Without any further constraints, the participants thus can use their own granularities and preferences to summarize the videos. Table(\ref{tab:stats}) exhibits that user have their own distinct preferences about the summary lengths.

\begin{table}[t]
	\small
			\centering
		\caption{\small{Some statistics about the lengths of the summaries generated by three annotators.}}
	\label{tab:stats}
	\begin{tabular}{ccccccccc}
	\toprule
		& User 1 & User 2 & User 3 & Oracle \\ \cmidrule{2-5}
		Min & 79 & 74 & 45 &  74\\
		Max & 174 & 222 & 352 & 200 \\
		Avg. & 105.75\scriptsize{$\pm$27.21}  & 133.33\scriptsize{$\pm$54.04}  & 177.92\scriptsize{$\pm$90.96} & 135.92\scriptsize{$\pm$45.99} \\ 
		\bottomrule
	\end{tabular}
	\vspace{-15pt}
\end{table}

\paragraph{Oracle Summaries.} Supervised video summarization approaches are conventionally trained on one target summary per video. Having obtained 3 user summaries per video, we aggregate them into one \textit{oracle summary} using a greedy algorithm that has been used in several previous works~\cite{gong2014diverse,sharghi2016query,sharghi2017query}, and train the model on them. We leave the details of the algorithm to the supplementary materials.  

\paragraph{Features.} We follow Zhang et al.~\cite{zhang2016video} in extracting the features using pre-trained GoogleNet~\cite{szegedy2015going}, after the pool5 layer, which results in a 1024-d feature representation for each shot in the video.


\begin{figure}[t]
    \centering
    \begin{subfigure}[b]{0.49\textwidth}
        \includegraphics[width=\textwidth]{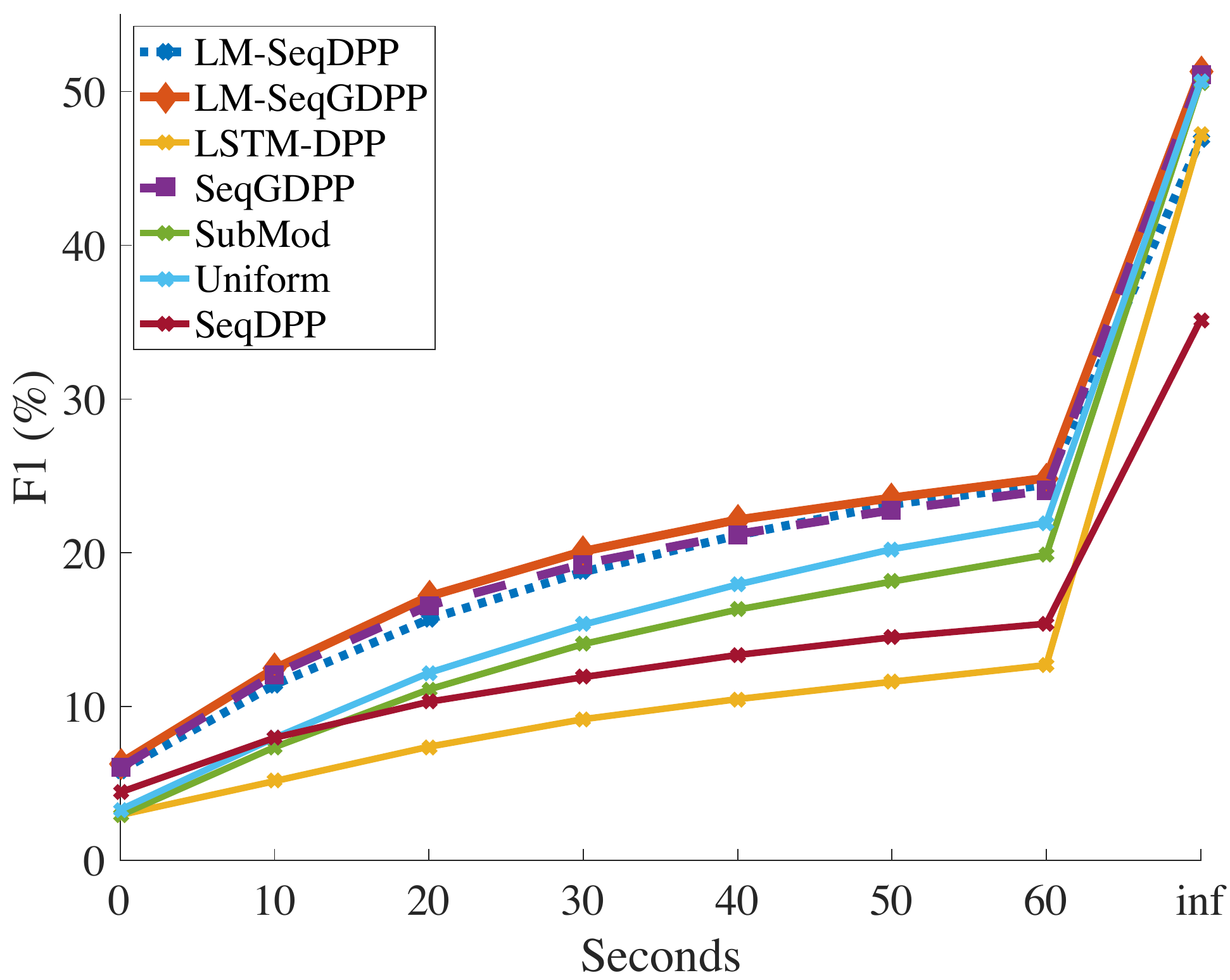}
        \caption{$\Pi$ temporal filter}
        \label{fig:g_pi}
    \end{subfigure}
    \begin{subfigure}[b]{0.49\textwidth}
        \includegraphics[width=\textwidth]{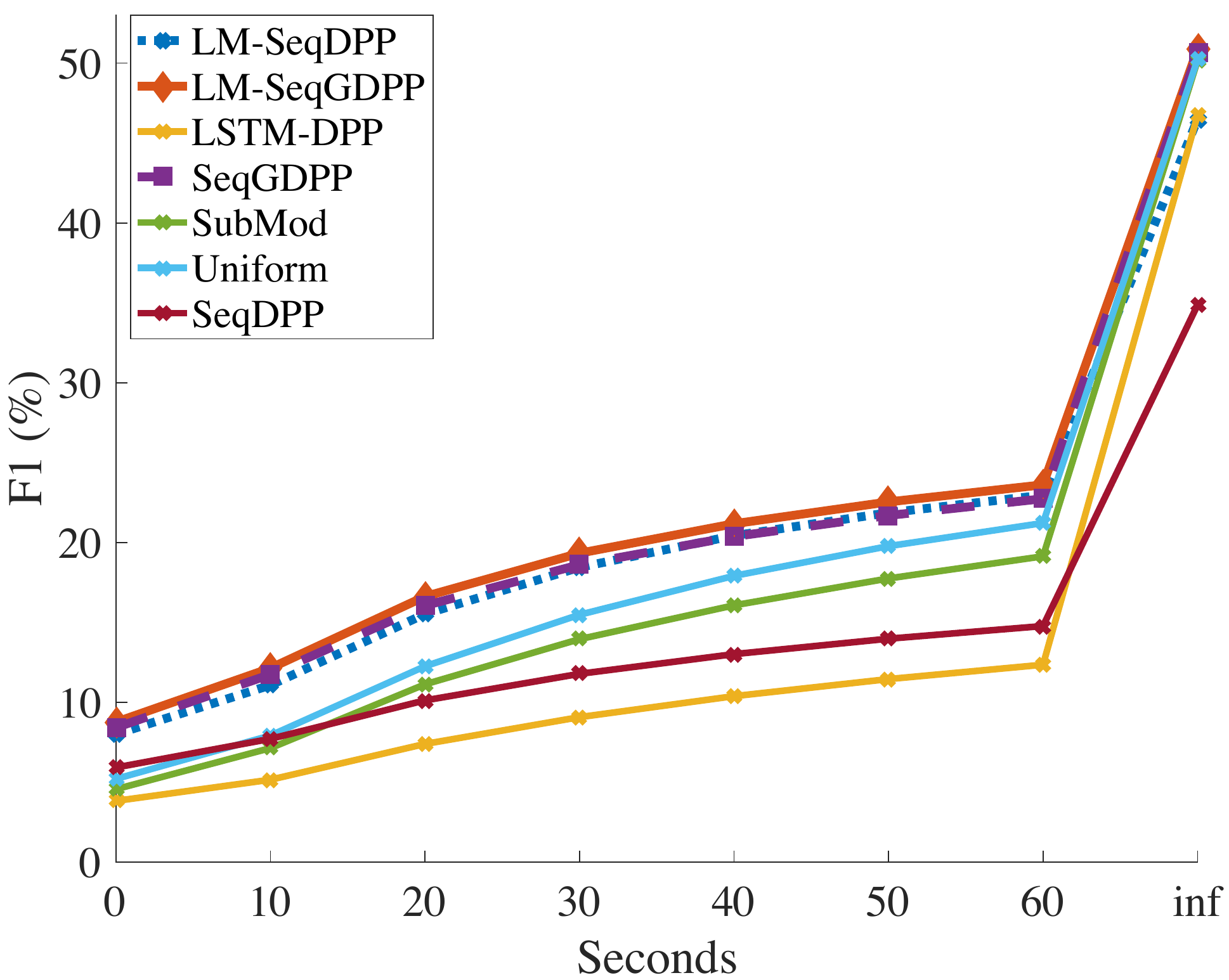}
        \caption{Gaussian temporal filter}
        \label{fig:g_g}
    \end{subfigure}
    \caption{Comparison results for generic video summarization task. x axis represent the temporal filter parameter. In case of $\Pi$ filter, it indicates how far a match can be temporally (in terms of seconds), whereas in the Gaussian filter, it is the kernel bandwidth.}\label{fig:generic}
\end{figure}
\paragraph{Evaluation.} \label{Eval}There has been a plethora of different approaches for evaluating the quality of video summaries including user studies~\cite{lee2015predicting,lu2013story}, using low-level or pixel-level measurements to compare system summaries versus human summaries~\cite{gong2014diverse,khosla2013large,kim2014joint,zhang2016summary,zhao2014quasi}, and temporal-overlap-based metrics defined for two summaries~\cite{gygli2014creating,gygli2015video,potapov2014category,zhang2016video}. We share the same opinion as~\cite{yeung2014videoset,sharghi2016query,sharghi2017query} in evaluating the summaries using high-level semantic information.

For measuring the quality of system summaries, Sharghi et al.~\cite{sharghi2017query} proposed to obtain dense shot-level concept annotation and convert them to semantic vectors where 1's and 0's indicate the presence or absence of a visual concepts such as \textsc{Sky, Car}, \textsc{Tree}, and etc. for that specific shot. It is straightforward to measure similarity between two shots using intersection-over-union (IoU) of their corresponding tags. For instance, if one shot is tagged by \{\textsc{Street,Tree,Sun}\} and the other by \{\textsc{Lady,Car,Street,Tree}\}, then the IoU is ${2}/{5} = 0.4$. Having defined the similarity measure between shots, one can conveniently perform maximum weight matching on the bipartite graph, where the user and system summaries are placed on opposing sides of the graph. 

\begin{figure}[t]
    \centering
    \includegraphics[width=\textwidth]{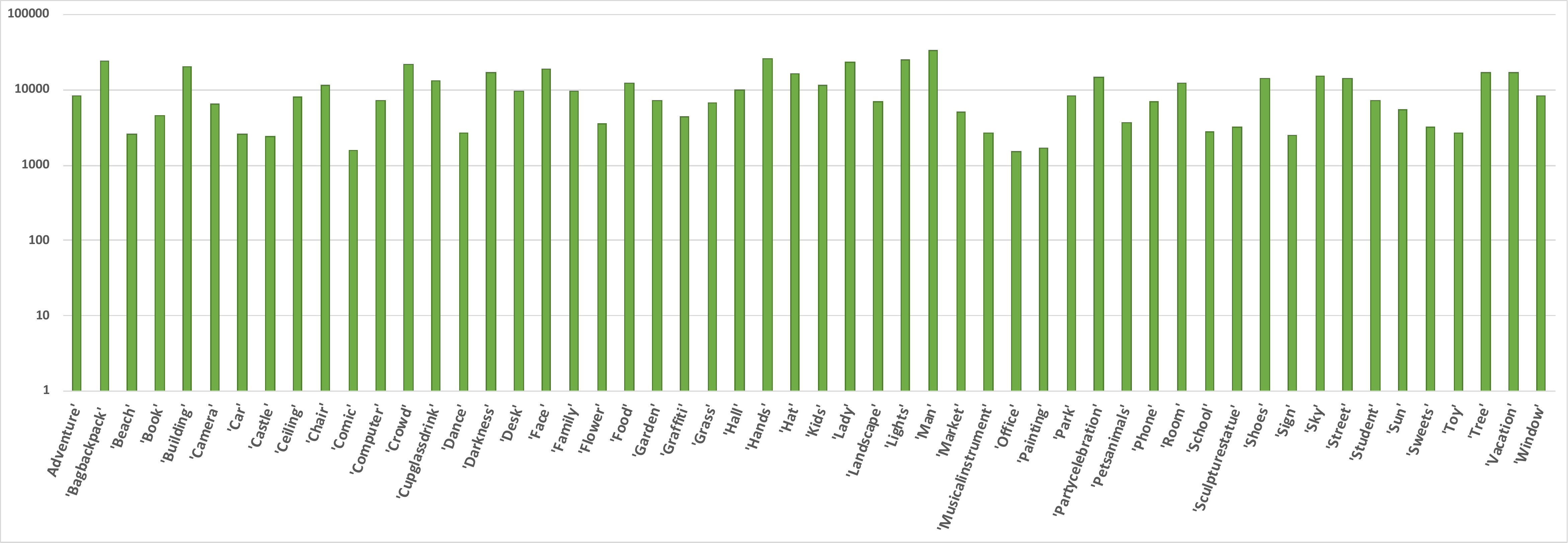}
        \caption{Count of concept appearances in the collected annotations accumulated over all 12 videos.}
        \label{fig:freq}
\end{figure}

To collect shot-level concept annotations, we start with the dictionary of~\cite{sharghi2017query}, and remove the concepts that do not appear often enough such as \textsc{Boat} and \textsc{Ocean} from it. Furthermore, we apply SentiBank detectors~\cite{borth2013large} (with over 1400 pre-trained classifiers) on the frames of the videos to make a list of visual concepts appearing commonly throughout the dataset. Next, by watching the videos, we select from this list the top candidates and append them into the final dictionary that includes 54 concepts. These steps are mandatory as our dataset contains over 3 times the video content in~\cite{sharghi2017query}. Figure~\ref{fig:freq} illustrates the appearance count of visual concept throughout our dataset. 

Having constructed a dictionary of concepts, we uniformly sample 5 frames from each shot and ask Amazon Mechanical Turk workers to tag them with the concepts. The instruction here is that a concept must be selected if it appears in any of the 5 frames. We hire 3 Turkers per shot and pool their annotations by taking the union. On average, each shot is tagged with $\sim$11 concepts. This is significantly larger than the average of 4 tags/shot in Sharghi et al.~\cite{sharghi2017query}, resulting in more reliable assessment upon evaluation.

While the metric introduced in~\cite{sharghi2017query} compares summaries using high-level concept similarities, it allows a shot in system summary to be matched with any shot in the user summary without any temporal restrictions. This causes at least two problems. First, for an important shot  in the gold summary, there is a chance we match it to a visually similar shot that may have happened long before or after. Second, since the shot similarities are necessarily positive, matching weakly similar shots that are temporally far, falsely increases the matching score. To fix these two issues, we modify this metric by applying a temporal filter on the measured similarities. We use two types of filters: 1) $\Pi$ (a.k.a rectangular) function and 2) Gaussian function. $\Pi$ filter sets the similarities outside of a time range to zero, hence forcing the metric to match a shot to its temporally close candidates. Gaussian filter on the other hand applies a decaying factor on farther matches. 

To evaluate a summary, we compare it to all 3 user-annotated summaries and average the scores. We report the performance by varying corresponding filter's parameters, the temporal window size and the bandwidth in $\Pi$ and Gaussian filters respectively, illustrated in Figure(\ref{fig:generic}). In addition, we compute the Area-Under-the-Curve (AUC) of average F1-scores in Table(\ref{tgresults}). It is worth mentioning that setting the parameters of the filters to infinity results in the same metric defined by Sharghi et al.~\cite{sharghi2017query}. Our metric is thus a generalization of the latter.

\paragraph{Data split.} In order to have a comprehensive assessment of the models, we employ leave-one-out strategy. Therefore, we run 12 set of experiments, each time leaving one video out for testing, two for validation (to tune hyper-parameters), and the remaining 9 for training the models. We report the average performance on all 12 videos later in this section.

\paragraph{Large-Margin Training/Inference.} Similar to practices in seq-2-seq learning~\cite{wiseman2016sequence,ranzato2015sequence}, we accelerate training by pre-training using standard sequential models, i.e. maximizing the likelihood of user summaries using SGD. This serves as a good network initialization, resulting in faster training process. At the test time, we follow Eq.(\ref{eInf}) to generate the system summary.

\paragraph{SeqGDPP Details.} Given the features that are extracted using GoogleNet, we compute Gaussian RBF kernels $\{\mL^{t(i)}\}_{i=1}^\cD$ over the video shots by varying the bandwidths $\sigma_i=1.2^k\sigma_0$, where $\sigma_0$ is the median of all pairwise distances between the video shots. Note that the base kernels $\{\mOmega^{t(i)}\}$ for {\gdpp}s and then computed through eq.~(\ref{eNewL}) such that they take account of the dependency between two adjacent time steps.

We also need to extract the feature vector $\phi(\sV_t)$ to capture the information in each video segment $\sV_t$. In eq.~(\ref{eMut}), we use such feature vector to help fine-tune the mean of the distribution $\vp$ over the possible subset sizes. Intuitively, larger subsets should be selected from segments with more frequent visual appearance changes. As such, we compute the standard deviation per feature dimension within the segment $\sV_t$ for  $\phi(\sV_t)$. 

There are three sets of parameters in {\ours}: $\alpha$ and $\vw$ in the distribution over the subset size, and $\{\beta_i\}$ for the convex combination of some base {\gdpp}s. We maximize the log-likelihood simply using gradient descent to solve for $\vw$ and $\{\beta_i\}$, and cross-validating $\alpha$.

\paragraph{Query-Focused Video Summarization.} As defined by Sharghi et al.~\cite{sharghi2016query}, due to the subjectivity of video summarization, it is appealing to personalize the summary based on user's preferences. Hence, in query-focused summarization, deciding whether to include a video shot in the summary or not, depends jointly on shot's relevance to a query term (that comes from the user) and its importance in the context of the video. In~\cite{sharghi2017query}, they made available a collection of 184 \{\textit{video},\textit{query}\} pair. To further assess our models, we compare them to the state-of-the-art query-focused video summarization frameworks in the supplementary material.

\begin{table}[t]
	\centering
	\small
	\caption{\small{Comparison results for supervised video summarization (\%). The AUCs are computed by the F1-score curves drawn in Figure~\ref{fig:generic} until the 60 seconds mark. The blue and red colors group the base model and its  large-margin version.}}
	\label{tgresults}
	\begin{tabular}{@{}lccc@{}}\toprule
		& \multicolumn{1}{c}{$\text{AUC}_{\Pi}$}& \phantom{abc}& \multicolumn{1}{c}{${\text{AUC}_\text{Gaussian}}$}\\ 
        \midrule
		Uniform                        & 12.33 && 12.36 \\
		SubMod~\cite{gygli2015video}   & 11.20 && 11.12 \\
		SuperFrames~\cite{gygli2014creating} & 11.46 && 11.28 \\
		LSTM-DPP~\cite{zhang2016video} & 7.38 && 7.36 \\ 
		SeqDPP~\cite{gong2014diverse}  & \textcolor{blue}{9.71} && \textcolor{blue}{9.56} \\
		\textbf{LM-SeqDPP}             & \textcolor{blue}{15.05} && \textcolor{blue}{14.69} \\
		\textbf{SeqGDPP}               & \textcolor{red}{15.29} && \textcolor{red}{14.86}\\
		\textbf{LM-SeqGDPP}            & \textcolor{red}{\textbf{15.87}} && \textcolor{red}{\textbf{15.43}}\\
		\bottomrule
	\end{tabular}
\end{table}

\subsection{Quantitative Results and Analyses}

In this section, we report quantitative results comparing our proposed models against various baselines:

-- \textit{Uniform}. As the name suggests, we sample shots with fixed step size from the video such that the generated summary has equal length (same number of shots) as the oracle summary.

-- \textit{SubMod}. Gygli et al.~\cite{gygli2015video} learn a convex combination of interestingness, representativeness, and uniformity from user summaries in a supervised manner. At the test time, given the expected summary length, that is the length of the oracle summary, model generates the summary.

-- \textit{SuperFrames}. In~\cite{gygli2014creating}, Gygli et al. first segment the video into superframes and then measure their individual importance scores. Given the scores, the subsets that achieve the highest accumulative scores are considered the desired summary. Since a shot is 5-second long  in our dataset, we skip the super-frame segmentation component. We  train a neural network consisting of three fully-connected layers to measure each shot's importance score, and then choose the subsets with the highest accumulated scores as the summary. 

-- \textit{LSTM-DPP}. In~\cite{zhang2016video}, Zhang et al. exploit LSTMs to model the temporal dependency between the shots of the video, and further use DPPs to enforce diversity in selecting important shots. Similar to previous baselines, this model has access to the expected summary length at the test time.

-- \textit{SeqDPP}. This is the original framework of Gong et al.~\cite{gong2014diverse}. Unlike other baselines, this model determines the summary length automatically.

Various interesting and promising observations can be made from Table(\ref{tgresults}) and Figure(\ref{fig:generic}):

1) Comparing SeqDPP and large-margin SeqDPP (regarded as LM-SeqDPP), we observe a significant performance boost. As illustrated in Figure(\ref{fig:generic}), the performance gap is consistently large throughout different filter parameters. Although both SeqDPP and LM-SeqDPP determine the summary length automatically, our speculations show that the latter makes summaries that resemble the oracle summaries in terms of both length and conveyed semantic information.

2) Comparing {\ours} to SeqDPP, for which users cannot tune the expected length of the summary, we can see that {\ours} significantly outperforms SeqDPP. This is not surprising since SeqDPP does not have a mechanism to take the user supplied summary length into account. As a result, the number of selected shots by SeqDPP is sometimes much less or more than the length of the user summary. 

3) Large-margin {\ours} (LM-SeqGDPP) performs slightly better than {\ours}, and it outperforms all models. As both models generate system summaries of equal length to the oracle, large-margin formulation helps making better summaries by optimizing for the evaluation metric.

4) As described earlier, our refined evaluation scheme is a generalization of the BM; by setting the filter parameters to infinity (hence no temporal restriction enforced by the filters), we can obtain the performance under the BM metric, represented by the last points of the curves in Figure(\ref{fig:generic}). While performance under our refined metric is significantly difference from model to model, under the BM metric, models perform almost the same. This is due to the problems we mentioned earlier in Section~\ref{Eval}, where we discussed the evaluation metric.

\section{Conclusion}


In this work, we made twofold contribution towards improving sequential determinantal point process-based models for supervised video summarization. We proposed a large-margin training scheme that facilitates learning models more effectively by addressing common problems in most seq2seq frameworks -- exposure bias and loss-evaluation mismatch. In modeling terms, we introduce a new probabilistic block \textit{G}DPP that when integrated into SeqDPP, the resulting model can accept priors about expected summary length. Furthermore, we compiled a large video summarization dataset consisting of 12 egocentric videos totalling over 60 hours content. Additionally, we collected 3 user-annotated summaries per video as well as dense concept annotations required for evaluation. Finally, we conduct experiments on the dataset to verify the effectiveness of the proposed models.



{\small
\bibliographystyle{splncs}
\bibliography{egbib}
}

\end{document}